\documentclass[10pt,twocolumn,letterpaper]{article}

\usepackage{cvpr}
\usepackage{times}
\usepackage{cite}
\usepackage{epsfig}
\usepackage{graphicx}

\usepackage{amsmath}
\usepackage{stfloats}
\usepackage[colorlinks,linkcolor=red]{hyperref}
\usepackage{amssymb}
\usepackage{algpseudocode}
\usepackage{amsmath}
\usepackage{picins}
\usepackage{stfloats}
\usepackage{algorithm}
\usepackage{algorithmicx}
\usepackage{algpseudocode}
\usepackage{amsmath}

\cvprfinalcopy 

\begin{document}

\title{One-Stage Inpainting with Bilateral Attention and Pyramid Filling Block }

\author{Hongyu Liu\\
\and
Bin Jiang\thanks{Corresponding author}\\
\and
Wei Huang \\
\and
Chao Yang\\
\and
College of Computer Science and Electronic Engineering\\
Hunan University\\
{\tt\small $\{$kumapower, jiangbin, hwei, yangchaoedu$\}$@hnu.edu.cn }
}

\maketitle

\begin{abstract}
 Recent deep learning based image inpainting methods which utilize contextual information and two-stage architecture have exhibited remarkable performance. However, the two-stage architecture is time-consuming, the contextual information lack high-level semantics and ignores both the semantic relevance and distance information of hole's feature patches, these limitations result in blurry textures and distorted structures of final result. Motivated by these observations, we propose a new deep generative model-based approach, which trains a shared network twice with different targets and utilizes a single network during the testing phase, so that we can effectively save inference time. Specifically, the targets of two training steps are structure reconstruction and texture generation respectively. During the second training, we first propose a Pyramid Filling Block (PF-block) to utilize the high-level features that the hole regions has been filled to guide the filling process of low-level features progressively, the missing content can be filled from deep to shallow in a pyramid fashion. Then, inspired by the classical bilateral filter~\cite{58}, we propose the Bilateral Attention layer (BA-layer) to optimize filled feature map, which synthesizes feature patches at each position by computing weighted sums of the surrounding feature patches, these weights are derived by considering both distance and value relationships between feature patches, thus making the visually plausible inpainting results. Finally, experiments on multiple publicly available datasets show the superior performance of our approach.
\end{abstract}

\section{Introduction}

Image inpainting is the task to generate the alternative structures and textures of a plausible hypothesis for missing regions in corrupted input images. It has a wide range of applications, such as distracting object removal, restoring damaged parts, etc. The core challenge of image inpainting is to construct global reasonable structures and generate local texture details. Some early patch-based works~\cite{1,2,3,18}attempt to fill missing holes with texture synthesis techniques, In~\cite{3}, Barnes et al. propose the Patch-Match algorithm which gradually fills in holes by searching for the best fitting patches from hole boundaries. These methods are promising in hallucinating detailed textures for background inpainting tasks. However, since these methods cannot capture high-level semantics, they struggle for reconstructing these locally unique patterns.

In contrast, some early deep convolution neural networks based approaches ~\cite{4,5,6,7} model the inpainting task as a conditional generation problem, which learns data distribution to capture the semantic information of the images. Although these methods can generate semantically plausible results, they fail to indistinguishably treat the structure and texture information, and they can not effectively utilize contextual information to generate the missing parts. Thus, they are limited in handling irregular holes and more likely to generate inpainting results with noise patterns or texture artifacts.

To deal with these problems, some recent researches ~\cite{9,38,46} incorporate the patch-based idea into deep convolutional networks to fill hole regions by replacing the hole feature with contextual features. These methods can ensure the global semantic consistency. However, they ignore the semantic relevance and feature continuity of generated contents, which is crucial for the local pixel continuity in the image level. Meanwhile, the contextual information are obtained from low-level feature maps or a single feature map, which is lack of high-level semantic information. Some methods~\cite{33,47} utilize two-stage architecture by recovering missing structures in the first stage and generating the final results using the reconstructed information in the second stage, these methods can make the generated contents more realistic, however, they require numerous computational resources. Other methods combine the above two ideas ~\cite{8,10,39,48} and include both advantages and disadvantages of the these two ideas. In ~\cite{45}, Liu et al. propose a coherent semantic attention layer with an iterative process to guarantee local feature coherency. However, this method is time-consuming since it utilizes the two-stage architecture and also lack of high-level contextual information in the attention operation. Meanwhile, iterative operation for reconstructing each patch ignores the relationship between the generated patch and its upper or lower location patches. In ~\cite{49}, Sagong et al. simplify the two-stage network structure to a single-stage encoder-decoder structure with a contextual attention module ~\cite{8} and parallel decoders to reduce computational resources. However, the contextual attention module will cause the semantic chasm that I mentioned above.


To overcome the above limitations, we train a shared network under the U-Net architecture twice with two different targets. Similar to the two-stage model ~\cite{48}, the first training is to recover meaningful structures and the second training is to generate textures. During the second training, we initialize the encoder and decoder with the weights obtained in the first training to generate textures, and we propose a PF-block which utilizes high-level contextual information to fill the hole regions of low-level encoder feature maps progressively from bottle-neck layer to up. The fully filled encoder feature maps are concatenated with the corresponding decoder feature by skip connection. The PF-block can avoid the transmission of invalid information in hole regions of encoder feature maps when using skip connection, and make the model perceive high-level contextual information. For further improving the details and guaranteeing feature coherency, the selected filled feature map will be optimized with SE-block ~\cite{56} in channel dimension first and reconstructed by a proposed BA-layer in spatial dimension second. The BA-layer can ensure the local correlation and long-term continuity of features. Specifically, the BA-layer is a revision of the classical bilateral filter ~\cite{58} in computer vision, which reconstructs feature patch at each position by computing weighted sums of the surrounding feature patches centered on this position, the weights are configured based on the distance and value similarity between the surrounding feature patches and the current reconstructed feature patch. Furthermore, there is an intersection between the surrounding areas of each feature patch, so these overlapping areas can model long-term continuity of features.

Qualitative and quantitative experiments are conducted on standard datasets CelebA ~\cite{13}, Places2 ~\cite{14}, and Paris StreetView ~\cite{40}. The experiments results demonstrate that our method can generate higher-quality inpainting results and consume less time than several state-of-the-art methods.

Our contributions are summarized as follows:
\begin{itemize}
\item We propose a novel bilateral attention layer that characterizes the value and distance relationship between deep feature patches to ensure local correlation and long-term continuity.
\item We propose a novel pyramid filling block to fill the hole regions of deep features progressively by using high-level contextual semantic features.
\item We design a training strategy to achieve the performance of two-stage architecture, so that reduces inference time during the testing phase. Meanwhile, experiments on multiple public datasets show that our method performs favorably in generating fine-detailed textures and visually plausible results.
\end{itemize}
\section{Related Works}
\begin{figure*}[t]
\centering
\includegraphics[scale=0.5]{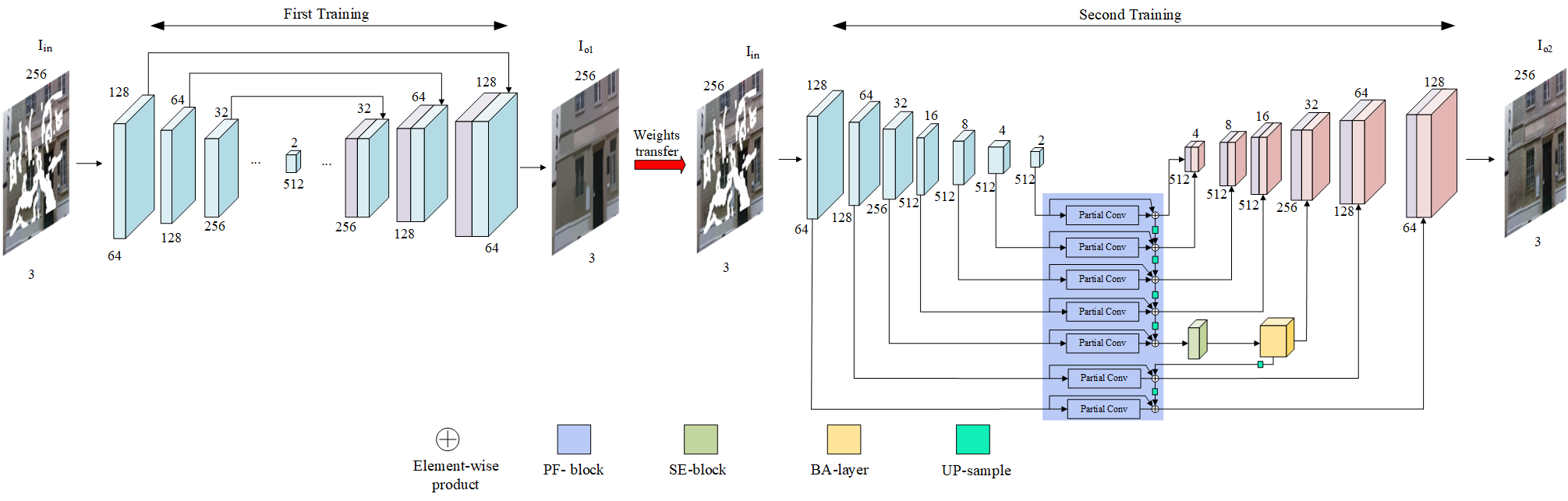}
\caption{The architecture of our model. The structure of encoder and decoder are same in two training steps. }
\label{img1}
\end{figure*}
Image inpainting approaches can be roughly divided into two categories: Non-learning inpainting approaches and Learning inpainting approaches. The Non-learning methods use a diffusion-based or patch-based strategy. The Learning methods learn the semantics of image to fulfill the inpainting task and generally train deep convolutional neural networks to infer the content of the missing regions.
\subsection{Non-learning inpainting}	
Non-learning methods with diffusion-based techniques such as ~\cite{18,50,51} propagate the neighborhood appearance information to the missing regions. However, they only consider surrounding pixels of missing regions, which can only deal with small holes in background inpainting tasks and may fail to generate meaningful structures. In contrast, the methods with the patch-based techniques ~\cite{20,28,3,25,52} fill missing regions by searching and copying similar and relevant patches from the exterior to the interior. These patch-based approaches perform well for small holes in background inpainting task, whereas if the missing regions of the image are large or the image contains rich semantic information, it tends to get a worse result.
\subsection{Learning inpainting}
Recently, Learning inpainting approaches often use deep learning based methods and GAN strategy to model the inpainting task as a conditional generation problem. Pathak et al. ~\cite{7} firstly introduce adversarial training ~\cite{41} to inpainting, albeit producing relatively low-resolution hallucinations. Iizuka et al.~\cite{4} propose local and global discriminators, assisted by dilated convolution ~\cite{59} to improve the inpainting quality and to handle rectangular masks at any location. However, it requires the previous processing steps to enforce the color coherency near the hole boundaries.

Multi-stage methods have also been investigated to ease the difficulty of training deep inpainting networks and get better predictions. Nazeri et al. ~\cite{47} propose a two-stage model EdgeConnect which first predicts salient edges and then generates inpainting results guided by edges. Song et al. ~\cite{33} divide the inpainting process into two steps: utilizing a segmentation prediction network to predict the semantic segmentation labels, then using a segmentation guidance network to refine details in the missing region. Xiong et al. ~\cite{61} present foreground-aware inpainting, which involves three stages, i.e., contour detection, contour completion and image completion, for the disentanglement of structure inference and content hallucination. Although these methods can improve the texture details and restore reasonable global structure, but they are time-consuming since the they utilize the multi-stage architecture and the generated features lack of contextual information.

Some methods utilize the contextual features in known regions to fill hole regions, since the contextual features are valid information. Yan et al. ~\cite{9} speculate the relationship between the contextual regions in the encoder layer and the associated hole region in the decoder layer for better predictions. Contextual attention~\cite{8} and patch-swap~\cite{10} search for a collection of background patches with the highest similarity to the first stage prediction. Ren et al.~\cite{48} introduce structure-aware appearance flow, which considers build long-term relationships between unknown areas and contextual areas, these areas are based on structural prediction which is the output of the first stage. Liu et al.~\cite{38} address this inpainting task via exploiting the partial convolutional layer and achieve mask-update operation, which yields convolutional results only depend on the contextual regions. Following this work, Yu et al.~\cite{39} present gate convolution that learns a dynamic mask-update mechanism and combines with SN-PatchGAN discriminator to achieve better predictions. In spite of the above approaches can generate perceptually realistic result in visual, they often give rise to color inconsistency and pixel discontinuity on generated image since they ignore the feature coherency of hole regions.

Liu et al.~\cite{45} propose a two-stage model with coherent semantic attention, which not only considers the relationship between the known areas and the missing areas but also consider the feature coherency of hole regions. However, this method lack of high-level contextual information in the attention operation and is time-consuming. Sagong et al.~\cite{49} simplify the coarse-to-fine structure to a single-stage encoder-decoder structure with contextual attention module~\cite{8} to reduce computational resources. However, the contextual attention module will cause the semantic chasm and pixels discontinuity.
\section{Approach}
The framework of our proposed model is shown in Fig \ref{img1}. Our model consists of two training steps with a shared network. We first train a network with U-net architecture to construct structure, then the learned weights of encoder and decoder are utilized to initialize the network of second training step to generate texture.  The BA-layer, PF-block and SE-block are embedded in the second training and testing step.  Let $I_{gts}$  and $I_{gt}$ be the ground-truth images of the first training step and second training step respectively, $I_{in}$ be the input to the network, $I_{o1}$ be the output of first training step and $I_{o2}$ be the final result.
\subsection{First training: structural construction}

The input $I_{in}$ is a 3$\times$256$\times$256 image with irregular holes. We feed $I_{in}$ to the generative network to output structural prediction $I_{o1}$ during the first training step, and the label image is $I_{gts}$. As the results of edge-preserved smooth methods RTV ~\cite{54}  can well represent global structures and we take these results as the $I_{gts}$. The structure of the network is the same as the generative network in ~\cite{11} without the last stage, which is composed of $4\times4 $ convolutions with skip connections to concatenate the features from each layer of the encoder and the corresponding layer of the decoder.

\subsection{Second training: Texture generation}
We use the same input $I_{in}$ in the second training step to train the generative network and get the final result $I_{o2}$ with valid textures. We initialize the encoder and decoder with the parameters obtained from the first training, so that the generative network can generate textures and reconstruct structures at the same time in the testing step. Note that we can employ singe network during testing with this two-step training strategy, which substantially reduces the computational resources. Moreover, we embed the PF-block, BA-layer and SE-block in both testing and second training step to help the network get better predictions. The PF-block filled the encoder feature maps from deep to shallow. Then, for the feature map with the size of $256\times32\times32$, the SE-block optimizes it in the channel dimension and the BA-layer reconstructs it in spatial dimension.

\subsection{Pyramid Filling Block }
\begin{figure}[t]
\centering
\includegraphics[scale=0.24]{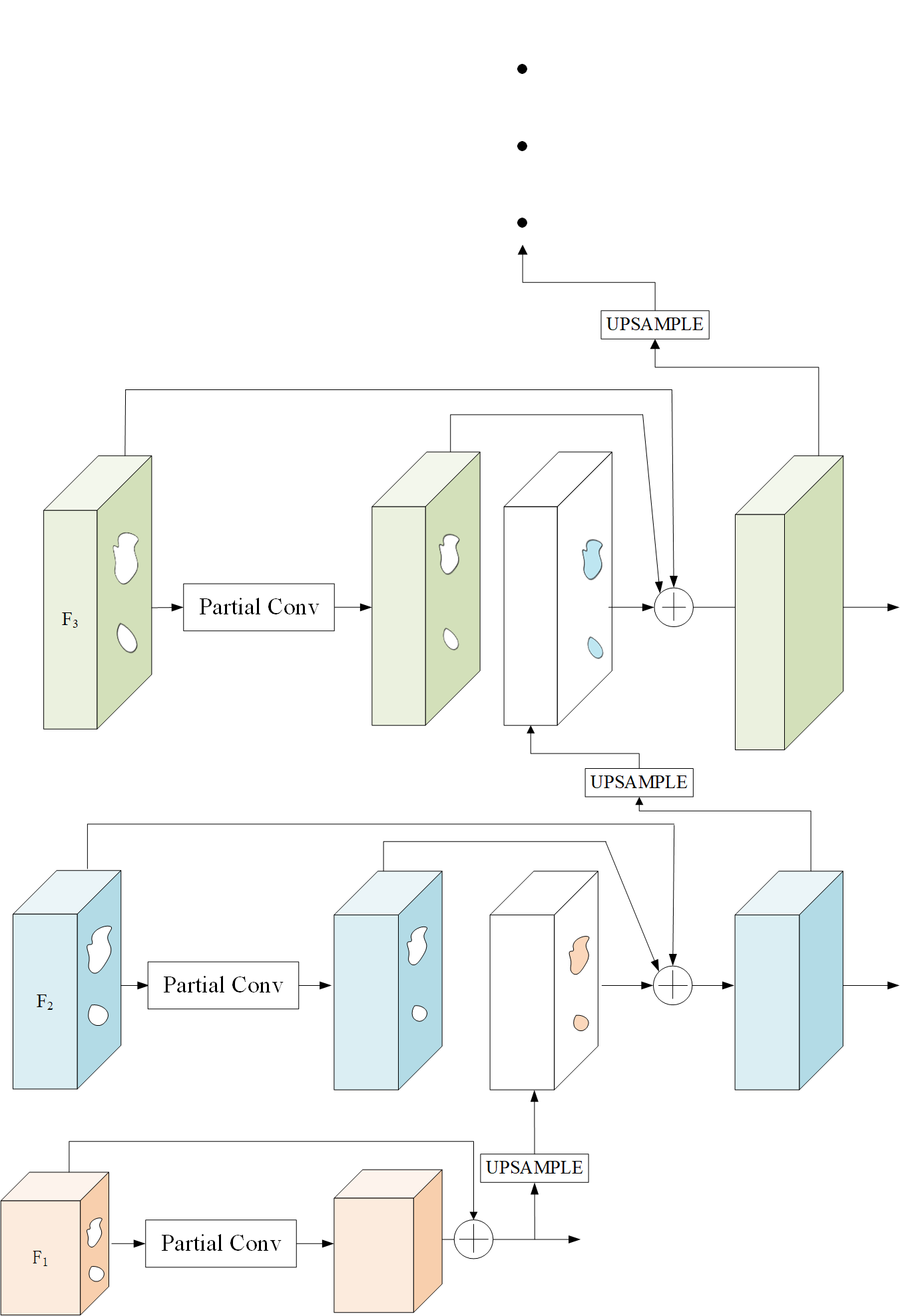}
\caption{The architecture of PF-block. $\oplus$ is element-wise sum, the irregular regions of each feature map denote hole regions. The $F_1$ with the size of $2\times2$ be filled by a single PConv first, then we use a progressive strategy to fill remaining hole areas of other feature maps from deep to shallow. The original and filled feature maps are added by the short connection.}
\label{img2}
\end{figure}
Some previous methods ~\cite{8,9,10} use the contextual areas of a single layer to fill in hole regions of this layer. However, since these methods are not aware of high-level semantics, the generated contents are always unreasonable. Liu et al. ~\cite{38} try to make the output conditioned only on the unmasked input (contextual regions) with partial convolution (PConv) to suppress color discrepancy and blurriness. However, the values of hole regions features are set to 0, which causes too much semantic information to be lost. Moreover, the feature information of each layer are from lower-level layer under the encoder-decoder architecture, which results in a lack of high-level semantics.

To handle these problems, we propose a PF-block which exploits the high level semantic features to guide the holes filling process progressively. Given a encoder of $L$ layers, we denote the feature maps from deep to shallow as $F_1$, $F_2$,...,$F_L$, the features constructed by PF-block in each layer from deep to shallow are denoted as:
\begin{equation}
\begin{gathered}
F_1=F_1 \oplus PConv(F_1)\\
F_2=F_2 \oplus PConv(F_2) \oplus (M\otimes f(F_1))\\
...,\\
F_L=F_L \oplus  PConv(F_L)\oplus (M\otimes f(F_{L-1}))\\
\end{gathered}
\label{eq1}
\end{equation}
Where the $\oplus$ is element-wise sum, the $\otimes $ is matrix multiplication, $f$ is up-sample operation and we utilize bilinear interpolation here. $M$ is the mask of each feature maps, it is a binarized matrix where 1 represents the missing region and 0 represents the background. So the hole regions of each feature map can be filled with high-level contextual features.

We take the last three layers $F_{1\sim 3}$ of encoder as an example, and Fig \ref{img2} illustrates the operation of the PF-block. Initially, a single PConv ~\cite{38} with the kernel size of $3$ conducts the process of filling the hole regions of each feature map. Except for the bottle-neck layer $F_1$ with the size of $2\times2$, the hole regions of other feature stages can not be filled with contextual patched by a single PConv. Thereafter, we use a progressive strategy to fill remaining hole areas of other feature maps from deep to shallow. Specifically, for the layer ${F_1}$ , we add the feature map which is reconstructed from the PConv to the original feature map with short connection as output to make the semantic information of hole areas more abundant, since the encoder has learned some meaningful parameters in the first training phase and the hole regions of original feature map are not all invalid. Then, for each of the other layer $F_{i(i\in (2\sim 3))}$, we up-sample the filled deeper feature map $F_{i+1}$ and add the features of hole regions to the output of PConv. Meanwhile, the output of $F_i$ will be combined with the original feature map with short connection similar to the process in $F_1$. During the progressive process, the hole regions of $F_{i+1}$ are fully filled first to help the $F_i$ fill the remaining hole areas after a single PConv, so that the $F_{i(i\in (2\sim 3))}$ can perceive high-level semantic information of $F_{i(i\in ((i-1)\sim 1))}$.
\subsection{Bilateral Attention}
\begin{figure}[t]
\centering
\includegraphics[scale=0.25]{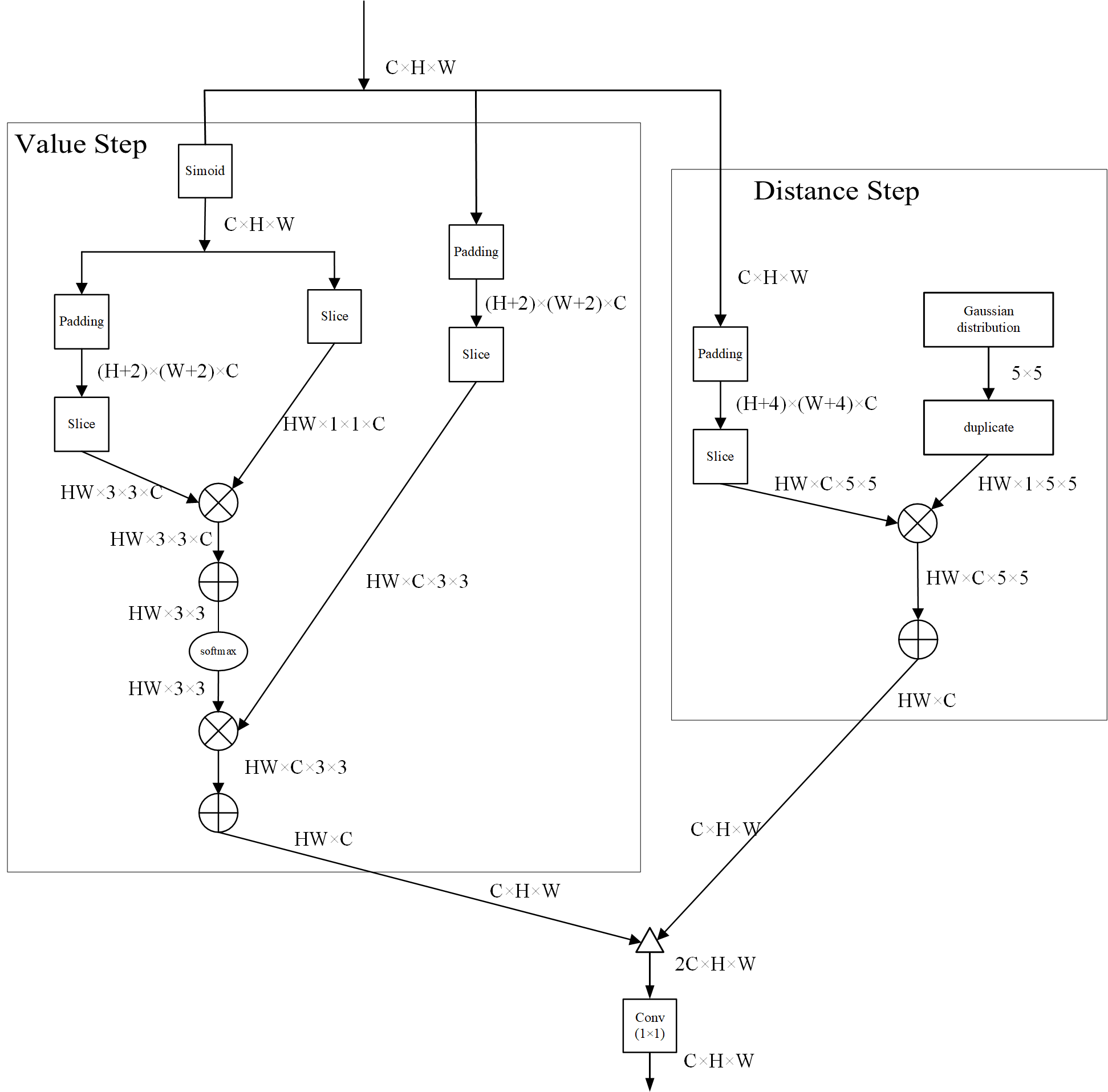}
\caption{The architecture of bilateral attention layer. The feature maps are shown by their dimensions, e.g. C$\times$H$\times$W.  $\otimes$ is broad cast matrix multiplication, $\otimes$ is element-wise addition in selected channel and $\bigtriangleup$ is concatenate operation. For two matrices with different dimensions, broadcast operations first broadcast features in each dimension to match the dimensions of the two matrices.}
\label{img3}
\end{figure}

The ~\cite{8,9,10,48} replace each patch inside the missing regions of a feature map with the patch on the contextual regions by attention or flow mechanism. However, they only fill the hole regions without considering the relationship between the hole features, which may result in lack of ductility and continuity of pixels in the final result. To handle this issue, Liu et al.~\cite{45} propose a coherent semantic attention (CSA) layer with an iterative method to model the relationship with adjacent generated patches and achieve good results. However, the CSA layer dose not properly construct the correlation between the current generated patches and the generated patches at its upper and lower positions. Moreover, the CSA layer is time-consuming and ignores the distance relationship of feature patches.

\subsubsection{Formulation}
To overcome the above limitations, we inspired by classical bilateral filter ~\cite{58} and propose the bilateral attention layer to consider the relationship between feature patches ($1\times1$) in two aspects: value and distance, so that we can make a more reasonable reconstruction of image features in spatial dimension. Since the image feature has been fully filled by PF-block, the BA-layer only needs to reconstruct feature map without necessary to fill hole regions like the above methods. The operation of bilateral attention is divided into two parallel steps: value and distance, and we define it in deep neural networks as:
\begin{equation}
\begin{gathered}
y_{distance_i}=\frac{1}{C(x)} \sum_{j(j\in s)}  g_{\alpha_s} (\lVert p_j-q_i \rVert) x_j \\
y_{value_i}=\frac{1}{C(x)} \sum_{j(j\in v)}  f(x_i,x_j) x_j\\
y_i = q (y_{distance_i}, y_{value_i})
\end{gathered}
\label{eq1}
\end{equation}
Here $i$ is the index of an feature patch position whose response is to be computed, $j$ is the index that enumerates the surrounding feature patch position which are centered on the position with the index of $i$. $x$ is the input feature, $y_{distance_i}$ and $y_{value_i}$ are the output features of distance step and value step respectively, $y$ is the final output feature of our BA-layer, and these three outputs are with the same size as $x$. Concretely, for the $y_{distance_i}$, we consider the relationship between feature patches from the aspect of distance to get it. The $s$ is a $5\times5$ square area which is centered on the position with the index of $i$, the $ g_{\alpha_s}$ is a Gaussian function to get the weight of each feature patch ($j$) in the square area, the $p$ and $q$ are the coordinate for feature patches. For the $y_{value_i}$, we calculate the value similarity between the feature pathes to get it, the pairwise function $f$ computes a scalar (representing relationship such as affinity) between $i$ and all $j$ in the surrounding area $v$ which is centered on the position with the index of $i$, the size of $v$ is $3\times3$. The output $y_{distance_i}$ and $y_{value_i}$ are normalized by a factor $C(x)$. Finally, the $y_{distance_i}$ and $y_{value_i}$ are concatenated together to get final output $y_i$.

During the value step, the pairwise function f can be defined as a dot-product similarity:
\begin{equation}\label{eqf}
f(x_i,x_j)= (x_i)^T(x_j)
\end{equation}
Similar to the Non-local block~\cite{56}, for a given i, $\frac{1}{C(x)}f(x_i,x_j)$  becomes the softmax computation along the dimension $j$.

During the distance step, the $ g_{\alpha_s}$ is a Gaussian function:
\begin{equation}\label{eqgus}
g_{\alpha_s} (\lVert p-q \rVert)= \frac{1}{2\pi\alpha_s^2}exp^{-((p_r-q_r)^2+(p_t-q_t)^2)/2\alpha_s^2}
\end{equation}
The $p$($p_r,p_t$) and $q$($q_r,q_t$) denote the coordinate for each position, $\alpha_s$ is smoothing parameters and we set $\alpha_s=1.5$ here. The $C(x)$ is $N$ and $N$ is the number of positions in region $s$.

Finally, for simplicity, the $y_{distance_i}$ and $y_{value_i}$ are spliced along the channel dimension, then halved the number of channels by a function $q$, the $q$ is implemented as a $1\times1$ convolution.

To compare with Non-local block~\cite{57} which uses the features of all positions in feature map to generate $y_i$ and only considers the value similarity between $x_i$ and $x_j$, the BA operation in Eq.\ref{eq1} is due to the fact that surrounding positions are considered, and expect for the value similarity, the distance between $x_i$ and $x_j$ is also considered as a aspect for calculating the weight of $x_j$. Since the inpainting is a generative task and the inputs are corrupted image, the response at a position is related to the surrounding positions, not all positions in the input feature maps.
Moreover, the surrounding areas for $x_i$ have an overlap area with the size of $3\times2$ in value step and $5\times2$ in position step respectively. The use of these overlapping regions allows each $y_i$ to contain the semantic information of the surrounding $y_i$, thereby ensuring long-term continuity of features.
\subsubsection{Implementation}
Fig \ref{img3} illustrates the implementation of the bilateral attention block, the pairwise computation in eql \ref{eqf} can be simply done by matrix multiplication as shown in Fig \ref{img3}. During the value step, we slice the input feature after the padding 0 operation and the weights of padding positions ($pp$) should be minimized, so we need do the sigmoid function in input feature first to ensure the dot-product result between $x_j(j\notin pp)$ and $x_i$ greater than 0. During the position step, the weights for $x_j$ of each $x_i$ are determined by the Gaussian distribution in eql \ref{eqgus}, and the weights for padding positions are set to 0.

\section{Loss Functions}
For better recovery of texture details and global structure, we incorporate pixel reconstruction loss and Relativistic Average LS adversarial loss ~\cite{12} to train our model in both training step. During the second training step, we add perceptual loss and style loss as constraints.

\textbf{Pixel Reconstruction Loss.} We adopt the $L_1$-norm error of the output image of both training steps as the pixel reconstruction loss:
\begin{equation}\label{eq:O2 Reconstruct Loss}
\begin{aligned}
L_{re1} = \lVert I_{o1}-I_{gts} \rVert_1
\end{aligned}
\end{equation}
\begin{equation}\label{eq:O1 Reconstruct Loss}
\begin{aligned}
L_{re2} =\lVert I_{o2}-I_{gt} \rVert_1
\end{aligned}
\end{equation}

\textbf{Perceptual Loss.}
To capture the high-level semantics and simulate human perception of images quality, we introduce the perceptual loss $L_perc$ defined on the  ImageNet-pretrained VGG-16.
\begin{equation}\label{eq:DR}
\begin{aligned}
L_{prec}=\mathbb{E}\Big \lbrack \sum_i \frac{1}{N_i} \lVert\Phi_{i}(I_{o2})-\Phi_{i} (I_{gt}) \rVert_1 \Big \rbrack
\end{aligned}
\end{equation}
where $ \Phi_{i}$ is the activation map of the i'th layer of VGG-16. For our work, $\Phi_{i}$ corresponds to activation maps from layers $relu1_-1$, $relu2_-1$, $relu3_-1$, $relu4_-1$ and $relu5_-1$.

\textbf{Style Loss.}
 Since our decoder consists of transpose convolution layers, we choose to use style loss as it was shown by Sajjadi et al. ~\cite{55} to be an effective tool to combat "checkerboard" artifacts caused by transpose convolution layers. Given feature maps of size $C_j \times H_j \times W_j$, style loss is computed by
 \begin{equation}\label{eq:DR}
\begin{aligned}
L_{style}=\mathbb{E}_j \Big \lbrack  \lVert G_j^\Phi (I_{o2})- G_j^\Phi (I_{gt}) \rVert_1 \Big \rbrack
\end{aligned}
\end{equation}
Where $ G_j^\Phi$ is a $C_j \times C_j$  Gram matrix constructed from activation maps $\Phi_j$. These activation maps are the same as those used in
perceptual loss.

\textbf{ Relativistic Average LS Adversarial Loss. }
Similar to ~\cite{45}, we use feature patch discriminator and patch discriminator to make our model synthesize more meaningful high-frequency details in both training stages, the Relativistic Average LS adversarial loss is adopted for our discriminators. The GAN loss term $L_G$ for generative network and the loss function $L_D$ for the discriminators are defined as:

\begin{equation}\label{eq:LR}
\begin{aligned}
L_R=-\mathbb{E}_{I_{r}}[D(I_{r},I_{f})^2]-\mathbb{E}_{I_{f}}[(1-D(I_{f},I_{r}))^2]
\end{aligned}
\end{equation}
\begin{equation}\label{eq:LD}
\begin{aligned}
L_D=-\mathbb{E}_{I_{r}}[(1-D(I_{r},I_{f}))^2]-\mathbb{E}_{I_{f}}[D(I_{f},I_{r})^2]
\end{aligned}
\end{equation}
Where $\mathbb{E}_{{I_{r}/I_f}}$ [.] represents the operation of taking average for all real/fake data in the mini-batch. In the first training step, the real image and fake image are $I_{gts}$ and $I_{o1}$ respectively. In the second training step, the real image and fake image are $I_{gt}$ and $I_{o2}$ respectively.

\textbf{ Model Objective.}
Taking the above loss functions into account, the overall objective of our model is defined as:
\begin{equation}\label{eq:final loss}
\begin{gathered}
L_{first} = \lambda_r L_{re1}+ \lambda_dD_R \\
L_{second} = \lambda_r L_{re2}+ \lambda_pL_{prec}+\lambda_sL_{style}+\lambda_dD_R
\end{gathered}
\end{equation}
Where the $L_{first}$ and $L_{second}$ are the overall objective of our model in first and second training steps respectively, $\lambda_r$, $\lambda_c$, $\lambda_s$ and $\lambda_d$ are the tradeoff parameters. In our implementation, we empirically set $\lambda_r=1$, $\lambda_p=0.1$, $\lambda_s=250$ and $\lambda_d=0.2$ .

\section{Experiments}
\begin{figure*}[t]
\centering
\includegraphics[scale=0.285]{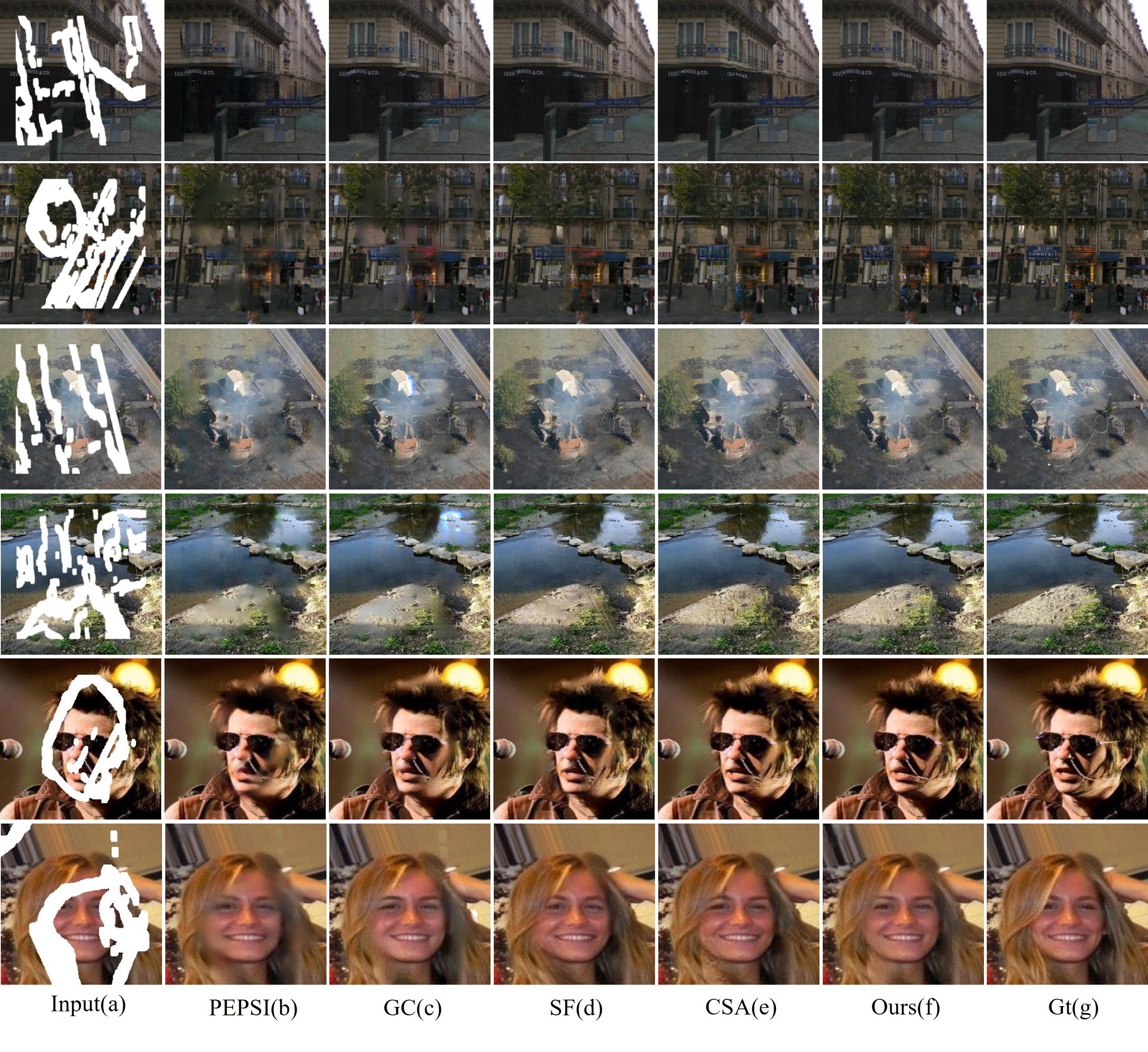}
\caption{ Qualitative comparisons on Places2, Celeba and Paris StreetView datasets. Comparison with PEPSI ~\cite{49}, SF ~\cite{48}, GC ~\cite{39}, CSA ~\cite{45} and Ours. First and second rows are results of Paris StreetView,  third and fourth are results of Places, fifth and sixth are results of Celeba.}
\label{img4}
\end{figure*}
We evaluate our method on three datasets: Places2~\cite{13}, CelebA~\cite{14}, and Paris StreetView~\cite{40}. We use the original training, testing, and validation splits for these three datasets. Data augmentation such as flipping is also adopted during training. Our model is optimized by the Adam algorithm~\cite{42} with a learning rate of $2\times10^{-4}$ and $\beta_1$ = 0.5. We train on a single NVIDIA 1080TI GPU (11GB) with a batch size of 1. The first training of CelebA model, Paris StreetView model, Place2 model are stopped after 6 epochs, 30 epochs and 60 epochs respectively.
The epochs of second training on three datasets are half of the first training. We compare our method with four state-of-the-art methods including PEPSI~\cite{49}, CSA~\cite{45}, GConv~\cite{39} and SF~\cite{48}. To fairly evaluate, we conduct experiments on irregular holes for our model and comparison methods. We obtain irregular masks from the work of PConv~\cite{38}. These masks are classified based on different hole-to-image area ratios (e.g., 0-10(\%), 10-20(\%), etc.)  All the masks and images for training and testing are with the size of 256$\times$256. The training is stopped after 30 epochs.

 \subsection{Qualitative Comparison }
\begin{figure*}[t]
\centering
\includegraphics[scale=0.25]{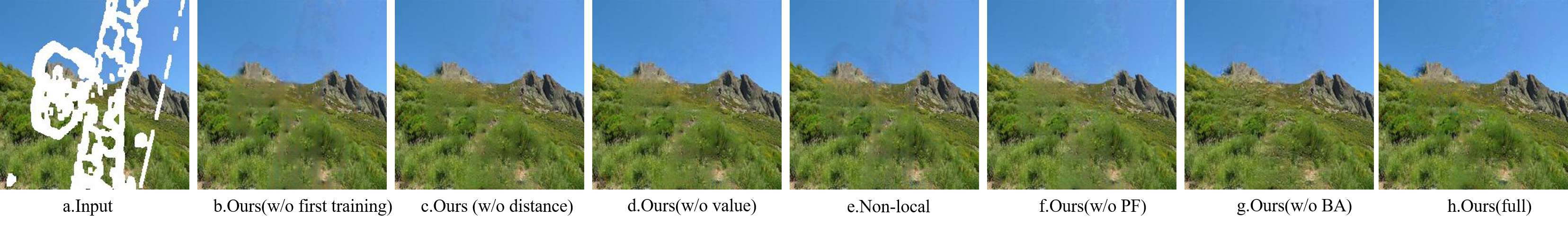}
\caption{ Visualization of ablation studies on Places2.}
\label{img5}
\end{figure*}

 Fig \ref{img4} shows the results on three datasets by our model and the competing methods. PEPSI ~\cite{49} is effective in semantic inpainting, but the results present distorted structure and blurry results. GConv~\cite{39} performances better due to the proposed gated convolution, but the process of mask update leads to unstable training and white spots. Moreover, these two methods have a phenomenon of semantic chasm since they utilize the contextual attention layer ~\cite{8} but lack of understanding of high-level semantic information. SF ~\cite{48} can get plausible result, but the continuities in color and rows do not hold well, this is mainly due to the fact that the flow operation does not consider the correlations between the deep features in hole regions. The CSA~\cite{45} performs well in constructing the pixel continuity, but the relationship between the pixels in the upper and lower positions is still not well modeled. In comparison to these competing methods, our model performs well in generating visually more plausible results with fine-detailed, and realistic textures.
 \subsection{Quantitative comparisons }

we conduct quantitative comparisons on Paris StreetView~\cite{40} with several mask ratios. We use common evaluation metrics,i.e., mean $L_1$ loss, mean $L_2$ loss, PSNR and SSIM to quantify the performance of the models. Moreover, our one-stage architecture aims to reduce time consumption, so the inference time is also an important evaluation metrics. Table \ref{tab1} lists the evaluation results, it can be seen that our method outperforms all the other methods except for the CSA ~\cite{45}. We think the reason for the CSA ~\cite{45} performs better than our model in some cases is that the two-stage architecture, but the time consumption for CSA ~\cite{45} is $29$ times larger than our model.

\begin{table}[t]
\centering
\setlength{\tabcolsep}{1.0mm}{
 \begin{tabular}{|l|c|c|c|c|c|c|c|}
 \hline

         & Mask         & PEPSI         & GC       & SF         & CSA         & Ours     \\ \hline
          &10-20\%      & 1.59       & 1.72         &1.88       &1.55          &\textbf{1.48} \\
L$_1^-$(\%) &20-30\%    & 2.03       & 2.23       & 2.36        &\textbf{1.93} &1.90          \\
         &30-40\%       & 3.80       & 3.65     & 3.71         &3.40           &\textbf{3.36}    \\
          &40-50\%      & 4.32      & 4.99      & 4.78         &4.57           &\textbf{4.36}    \\\hline
          &10-20\%      & 0.16       & 0.19     &0.16          &0.12            &\textbf{0.11}   \\
L$_2^-$(\%)&20-30\%     & 0.26       & 0.35    & 0.24         &0.19   &\textbf{0.19} \\
         &30-40\%       & 0.45       & 0.69    & 0.60         &0.55           &\textbf{0.52}    \\
          &40-50\%      & 0.77       & 0.96     & 0.84        &0.82            &\textbf{0.75}     \\\hline

          &10-20\%      & 29.21     & 27.16    &29.18   &\textbf{30.65} &30.60  \\
PSNR$^+$  &20-30\%      & 26.92      & 24.98    & 27.31 &\textbf{28.64} &28.51 \\
         &30-40\%       & 22.69     &21.56      & 23.01  &23.46         &\textbf{23.60}    \\
          &40-50\%      & 21.26     & 20.06     & 21.57   &21.78        &\textbf{22.08}    \\\hline

          &10-20\%      & 0.929     &  0.926     &0.952  &\textbf{0.966} &0.964  \\
SSIM$^+$  &20-30\%      & 0.878    & 0.885     & 0.921 &\textbf{0.941} &0.936 \\
         &30-40\%       & 0.823    & 0.802     & 0.832  &0.846 &\textbf{0.852}    \\
          &40-50\%      & 0.670    & 0.678     & 0.706  &0.719 &\textbf{0.720}     \\\hline

Time$^-$(ms) &-      & 18.9     & 27.3     & 36.4  &261.6 &\textbf{8.8}     \\\hline
 \end{tabular}}
 \caption{Comparison results over Paris StreetView between PEPSI ~\cite{49}, SF ~\cite{48},GConv ~\cite{39}, CSA ~\cite{45} and Ours. $^-$Lower is better. $^+$Higher is better}
\label{tab1}
\end{table}

\subsection{Ablation Studies}
\begin{figure}[h]
\centering
\includegraphics[scale=0.18]{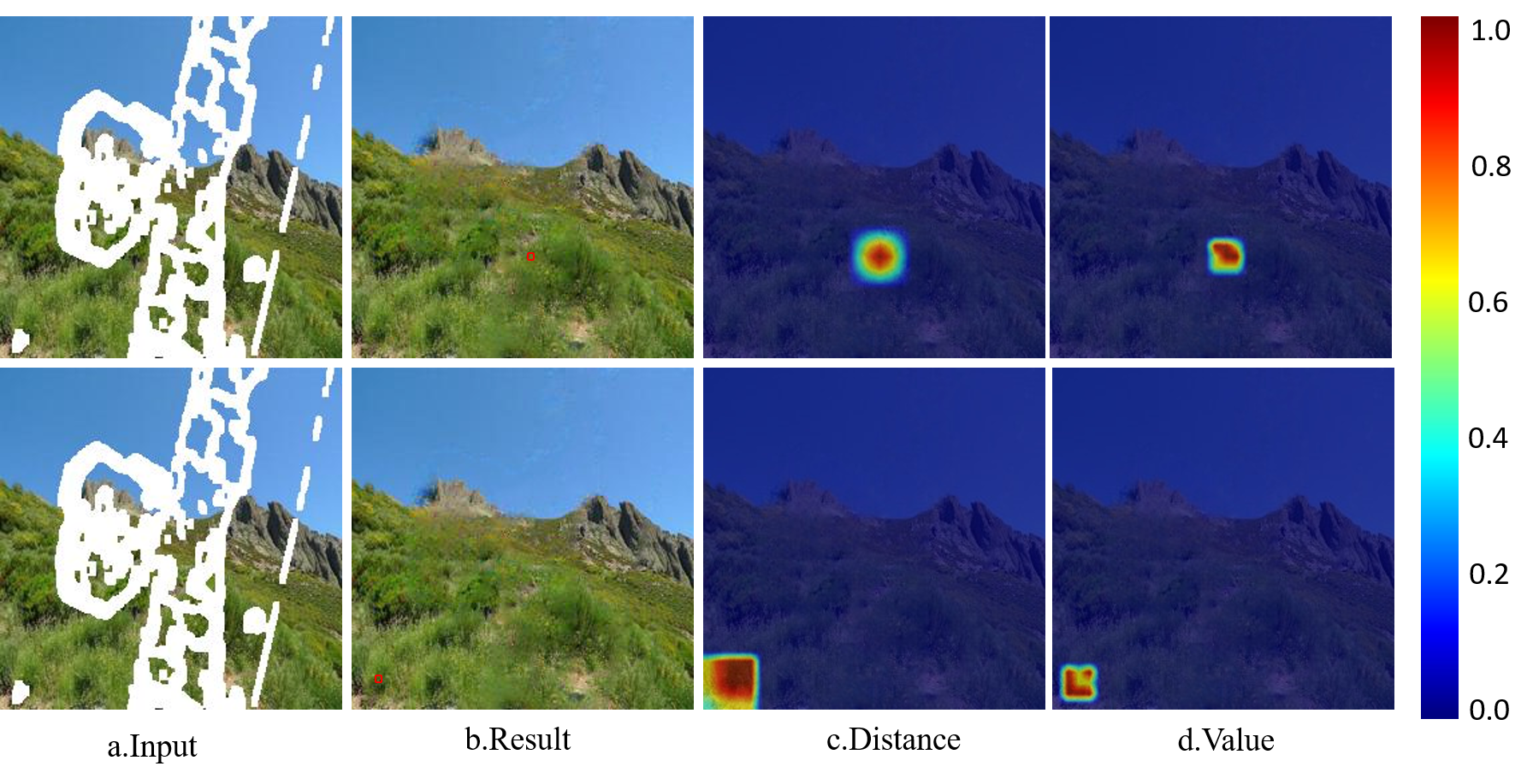}
\caption{ The visualization of attention map for two pixels. (a) is input, (b) is result of our model, (c) and (d) are attention maps of distance and value aspects respectively. The red square in (b) marks the position of the pixel.}
\label{img6}
\end{figure}
\begin{table}[t]
\small
\centering
\setlength{\tabcolsep}{0.16mm}{
 \begin{tabular}{|l|c|c|c|c|c|}
 \hline

  Methods      & 10-20\%         & 20-30\%        & 30-40\%        & 40-50\%       \\ \hline
 w/o first  & 30.25/0.949       &28.16/0.913         &24.90/0.848           &22.84/0.718                  \\
 w/o distance   & 31.12/0.951        & 28.87/0.921       & 25.15/0.852          &23.16/0.725        \\
  w/o value       & 31.23/0.956        & 28.96/0.921       & 25.36/0.854          &23.14/0.727    \\                                                                            w/o BA     & 31.14/0.954       & 28.64/0.912      & 25.22/0.850             &23.01/0.719                \\
 w/o PF       & 31.39/0.956       & 29.01/0.922      & 25.43/0.856          & 23.16/0.728                \\
 Non-local         & 31.38/0.956      & 28.89/0.92     &25.36/0.854                 &23.15/0.727             \\ \hline
 full                & \textbf{31.74}/\textbf{0.959}       & \textbf{29.27}/\textbf{0.927}    & \textbf{25.87}/\textbf{0.863}               &\textbf{23.44}/\textbf{0.739}              \\ \hline

 \end{tabular}}
 \caption{Ablation studies (PSNR/SSIM) on Places2.}
\label{tab2}
\end{table}
We conduct experiments to compare the performance of several variants of our model on Place2 for ablation studies, the quantitative and qualitative results are shown in Table \ref{tab2} and Fig \ref{img5} respectively.

\textbf{Effect of BA-layer}
Our BA-layer consists of value step and distance step, so we train our model without value step (w/o value), distance step (w/o distance) and BA-layer (w/o BA) respectively to make comparisons. As shown in Fig \ref{img5}, Ours (w/o value) and Ours (w/o distance) are smoothed excessively, Ours (w/o BA) generates unnatural texture. In contrast, Ours (full) with BA-layer weighs the smoothness of both value and distance aspects and ensure the local feature correlation and long-term pixel continuity (see Fig \ref{img5}(h)). Moreover, we replace the BA-layer with the Non-local block to make a comparison, as shown in Fig \ref{img5}(e), the inpainted result has some artifacts and noises since each feature patch contains too much irrelevant information. Compared with it, our BA-layer (full) only pays attention to the relationship between surrounding feature patches and central feature patch. Meanwhile, Ours (full) model also performances better than the above variants in PSNR and SSIM (see Table \ref{tab2}). We visualize the attention map of two pixels in Fig \ref{img6}, where the red square marks the position of the pixel of our inpainted results(see Fig \ref{img6} (b)), the distance and value attention maps are shown in Fig \ref{img6} (c) and (d) respectively.

\textbf{Effect of PF-block}
We train our model without (w/o) PF-block to make comparison, the BA-layer is embeded in original layer with resolution of $32\times32$. To compare with our full model, the qualitative results of model (w/o PF-block) seems more blurred (see Fig \ref{img5}(f)), the PSNR and SSIM of the model (w/o PF-block) are lower (see Table \ref{tab2}), especially in the case of large hole regions ($30\%-50\%$). This is because filling large holes requires a lot of semantic information and PF-block can provide this information to our full model.

\textbf{Effect of One-stage architecture}
We train our model without first training step (w/o first training) to evaluate the effect of our training strategy, the training configuration is the same as the second training in our paper. As shown in table \ref{tab2}, the metrics drop a lot when compare to our full model. Meanwhile, obvious blurriness and artifacts can be observed from Fig \ref{img5}(b).

\section{Conclusion}
This paper proposed a one-stage architecture with bilateral attention layer and pyramid filling block for image inpainting. The bilateral attention layer ensures the local correlation and long-term continuity of feature patches. Meanwhile, the pyramid fill block helps our model fill void regions with high-level semantic information to achieve better predictions. Moreover, the one-stage architecture is effective in reducing the time. Experiments have verified the effectiveness of our proposed methods.

{\small
 \bibliographystyle{ieee_fullname}
\bibliography{egbib}

\begin{thebibliography}{10}\itemsep=-1pt

\bibitem{2}
Efros~Alexei A and Leung~Thomas K.
\newblock Texture synthesis by nonparametric sampling.
\newblock {\em ICECCS}, 2001.

\bibitem{51}
C. Ballester, M. Bertalmio, V. Caselles, G. Sapiro, and J. Verdera.
\newblock Filling-in by joint interpolation of vector fields and gray levels.
\newblock {\em IEEE transactions on image processing}, 10.

\bibitem{3}
Connelly Barnes, Eli Shechtman, Adam Finkelstein, and Dan~B Goldman.
\newblock Patchmatch: A randomized correspondence algorithm forstructural image
  editing.
\newblock {\em ACM Transactions on Graphics}, 28, 2009.

\bibitem{18}
Marcelo Bertalmio, Guillermo Sapiro, Vincent Caselles, and Coloma Ballester.
\newblock Image inpainting.
\newblock {\em SIGGRAPH}, 2000.

\bibitem{40}
Doersch Carl, Singh Saurabh, Gupta Abhinav, Sivic Josef, and Efros~Alexei A.
\newblock What makes paris look like paris?
\newblock {\em ACM Transactions on graphics}, 31(4), 2012.

\bibitem{46}
Chao Li Ming-Ming Cheng Wangmeng Zuo Xiao Liu Shilei Wen Errui~Ding
  Chaohao~Xie, Shaohui~Liu.
\newblock Image inpainting with learnable bidirectional attention maps.
\newblock {\em ICCV}, 2019.

\bibitem{49}
Min cheol Sagong, Yong goo Shin, Seung wook Kim, Seung Park, and Sung jea Ko.
\newblock Pepsi : Fast image inpainting with parallel decoding network.
\newblock {\em CVPR}, 2019.

\bibitem{20}
Antonio Criminisi, Patrick P$\overline{e}$rez, and Kentaro Toyama.
\newblock Region filling and object removal by exemplar-based image inpainting.
\newblock {\em IEEE Transactions on image processing}, 13(9):1200--1212, 2004.

\bibitem{42}
Kingma Diederik and Ba Jimmy.
\newblock Adam: A method for stochastic optimization.
\newblock {\em ICLR}, 2015.

\bibitem{1}
Alexei~A Efros and William~T Freeman.
\newblock Image quilting for texture synthesis and transfer.
\newblock {\em SIGGRAPH}, 2001.

\bibitem{41}
Ian Goodfellow, Jean Pouget-Abadie, Mehdi Mirza, Bing Xu, David Warde-Farley,
  Sherjil Ozair, Aaron Courville, and Yoshua Bengio.
\newblock Generative adversarial networks.
\newblock {\em NIPS}, 2014.

\bibitem{56}
Jie Hu, Li Shen, and Gang Sun.
\newblock Squeeze-and-excitation networks.
\newblock {\em CVPR}, 2018.

\bibitem{4}
Satoshi Iizuka, Edgar Simo-Serra, and Hiroshi Ishikawa.
\newblock Globally and locally consistent image completion.
\newblock {\em ACM Transactions on Graphics}, 36(4), 2017.

\bibitem{11}
Phillip Isola, Jun-Yan Zhu, Tinghui Zhou, and Alexei~A Efros.
\newblock Image-to-image translation with conditional adversarial networks.
\newblock {\em CVPR}, 2017.

\bibitem{12}
Alexia Jolicoeur-Martineau.
\newblock The relativistic discriminator: a key element missing from standard
  gan.
\newblock {\em arXiv preprint arXiv: 1807.00734}, 2018.

\bibitem{50}
A. Levin, A. Zomet, and Y. Weiss.
\newblock Learning how to inpaint from global image statistics.
\newblock {\em IEEE}, 2003.

\bibitem{5}
Yijun Li, Sifei Liu, Jimei Yang, and Ming-Hsuan Yang.
\newblock Generative face completion.
\newblock {\em CVPR}, 2017.

\bibitem{38}
Guilin Liu, Fitsum~A Reda, Kevin~J Shih, Ting-Chun Wang, Andrew Tao, and Bryan
  Catanzaro.
\newblock Image inpainting for irregular holes using partial convolutions.
\newblock {\em ECCV}, 2018.

\bibitem{45}
Hongyu Liu, Bin Jiang, Yi Xiao, and Chao Yang.
\newblock Coherent semantic attention for image inpainting.
\newblock {\em ICCV}, 2019.

\bibitem{13}
Ziwei Liu, Ping Luo, Xiaogang Wang, and Xiaoou Tang.
\newblock Deep learning face attributes in the wild.
\newblock {\em ICCV}, 2015.

\bibitem{54}
LiXu, QiongYan, YangXia, and Jiaya Jia.
\newblock Structure extraction from texture via relative total variation.
\newblock {\em ACM Transactions on Graphics (TOG)}, 31.

\bibitem{47}
Kamyar Nazeri, Eric Ng, Tony Joseph, Faisal Qureshi, and Mehran Ebrahimi.
\newblock Edgeconnect: Generative image inpainting with adversarial edge
  learning.
\newblock {\em ICCVW}, 2019.

\bibitem{7}
Deepak Pathak, Philipp Krahenbuhl, Jeff Donahue, Trevor Darrell, and Alexei~A
  Efros.
\newblock Context encoders: Feature learning by inpainting.
\newblock {\em CVPR}, 2016.

\bibitem{6}
RaymondYeh, ChenChen, TeckYianLim, Mark Hasegawa-Johnson, and Minh N.Do.
\newblock Semantic image inpainting with perceptual and contextual losses.
\newblock {\em arXiv preprint arXiv:1607.07539}, 2016.

\bibitem{48}
Yurui Ren, Xiaoming Yu, Ruonan Zhang, Thomas~H. Li, Shan Liu, and Ge Li.
\newblock Structureflow: Image inpainting via structure-aware appearanceflow.
\newblock {\em ICCV}, 2019.

\bibitem{55}
M.~S.~M. Sajjadi, B. Scholkopf, and M. Hirsch.
\newblock Enhancenet: Single image super-resolution through automated texture
  synthesis.
\newblock {\em ICCV}, 2017.

\bibitem{28}
Darabi Soheil, Shechtman Eli, Barnes Connelly, Dan~B Goldman, and Sen Pradeep.
\newblock Image melding: Combining inconsistent images using patch-based
  synthesis.
\newblock {\em ACM Transactions on graphics}, 31(4), 2012.

\bibitem{10}
Yuhang Song, Chao Yang, Zhe Lin, Xiaofeng Liu, Qin Huang, Hao Li, and CC Jay.
\newblock Contextual-based image inpainting: Infer, match, and translate.
\newblock {\em ECCV}, 2018.

\bibitem{33}
Yuhang Song, Chao Yang, Yeji Shen, Peng Wang, Qin Huang, and C~C~Jay Kuo.
\newblock Spg-net: Segmentation prediction and guidance network for image
  inpainting.
\newblock {\em BMVC}, 2018.

\bibitem{58}
C. Tomasi and R. Manduchi.
\newblock Bilateral filtering for gray and color images.
\newblock {\em ICCV}, 1998.

\bibitem{57}
Xiaolong Wang, Ross Girshick, Abhinav Gupta, and Kaiming He.
\newblock Non-local neural networks.
\newblock {\em CVPR}, 2018.

\bibitem{61}
Wei Xiong, Jiahui Yu, Zhe Lin, Jimei Yang, Xin Lu, Connelly Barnes, and Jiebo
  Luo.
\newblock Foreground-aware image inpainting.
\newblock {\em CVPR}, 2019.

\bibitem{52}
Zongben Xu and Jian Sun.
\newblock Image inpainting by patch propagation using patch sparsity.
\newblock {\em IEEE Transactions on Image Processing (TIP)}, 1.

\bibitem{25}
Zongben Xu and Jian Sun.
\newblock Image inpainting by patch propagation using patch sparsity.
\newblock {\em IEEE transactions on image processing}, 19(5):1153--1165, 2010.

\bibitem{9}
Zhaoyi Yan, Xiaoming Li, Mu Li, Wangmeng Zuo, and Shiguang Shan.
\newblock Shift-net: Image inpainting via deep feature rearrangement.
\newblock {\em ECCV}, 2018.

\bibitem{59}
Fisher Yu and Vladlen Koltun.
\newblock Multi-scale context aggregation by dilated convolutions.
\newblock {\em arXiv preprint arXiv:1511.07122}, 2015.

\bibitem{8}
Jiahui Yu, Zhe Lin, Jimei Yang, Xiaohui Shen, Xin Lu, and Thomas~S Huang.
\newblock Generative image inpainting with contextual attention.
\newblock {\em CVPR}, 2018.

\bibitem{39}
Jiahui Yu, Zhe Lin, Jimei Yang, Xiaohui Shen, Xin Lu, and Thomas~S Huang.
\newblock Free-form image inpainting with gated convolution.
\newblock {\em ICCV}, 2019.

\bibitem{14}
Bolei Zhou, Agata Lapedriza, Aditya Khosla, Aude Oliva, and Antonio Torralba.
\newblock Places: A 10 million image database for scene recognition.
\newblock {\em IEEE Transactions on Pattern Analysis and Machine Intelligence},
  2017.

\end{thebibliography}
}

\end{document}